\documentclass[10pt,twocolumn,letterpaper]{article}

\usepackage{cvpr}
\usepackage{times}
\usepackage{epsfig}
\usepackage{graphicx}
\usepackage{amsmath}
\usepackage{amssymb}
\usepackage{booktabs}
\usepackage{subcaption}
\usepackage[resetlabels,labeled]{multibib}

\usepackage[breaklinks=true,bookmarks=false]{hyperref}

\cvprfinalcopy

\newcites{S}{References}
\begin{document}

\title{Actor and Action Video Segmentation from a Sentence}

\author{Kirill Gavrilyuk, Amir Ghodrati, Zhenyang Li, Cees G. M. Snoek\\
QUVA Lab, University of Amsterdam\\
{\tt\small \{kgavrilyuk, a.ghodrati, zhenyangli, cgmsnoek\}@uva.nl}
}

\maketitle

\begin{abstract}
This paper strives for pixel-level segmentation of actors and their actions in video content. Different from existing works, which all learn to segment from a fixed vocabulary of actor and action pairs, we infer the segmentation from a natural language input sentence. This allows to distinguish between fine-grained actors in the same super-category, identify actor and action instances, and segment pairs that are outside of the actor and action vocabulary. We propose a fully-convolutional model for pixel-level actor and action segmentation using an encoder-decoder architecture optimized for video. To show the potential of actor and action video segmentation from a sentence, we extend two popular actor and action datasets with more than 7,500 natural language descriptions. Experiments demonstrate the quality of the sentence-guided segmentations, the generalization ability of our model, and its advantage for traditional actor and action segmentation compared to the state-of-the-art.
\end{abstract}

\section{Introduction} \label{sec:intro}
\begin{figure}[t!]
    \centering 
    \includegraphics[width=\linewidth]{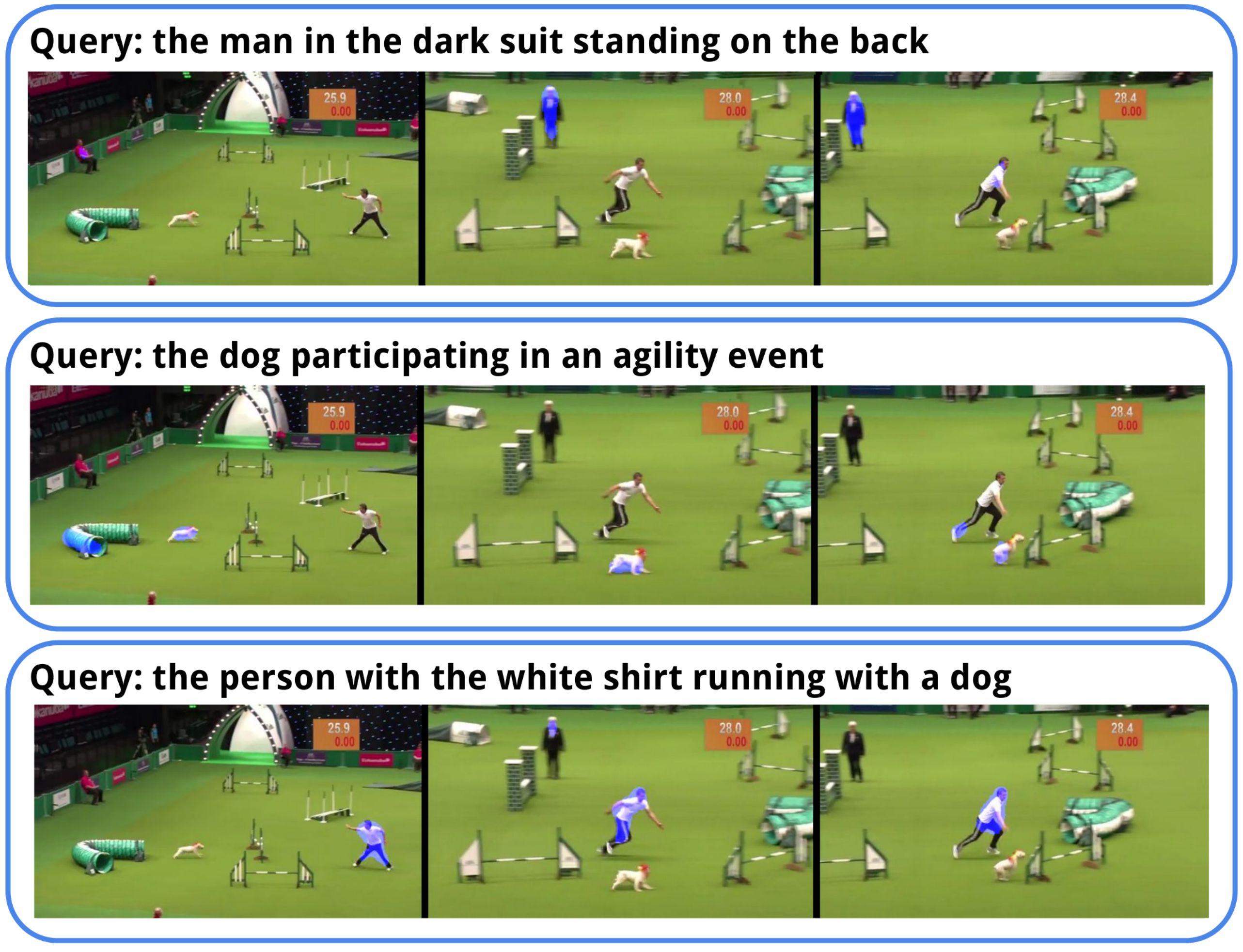} 
    \caption{From a natural language input sentence our proposed model generates a pixel-level segmentation of an actor and its action in video content.}
    \label{fig:intro}
\end{figure}

The goal of this paper is pixel-level segmentation of an actor and its action in video, be it a person that climbs, a car that jumps or a bird that flies. Xu \etal~\cite{xu2015fly} defined this challenging computer vision problem in an effort to lift video understanding beyond the more traditional work on spatio-temporal localization of human actions inside a tube, \eg \cite{MettesBMVC17, TianCVPR13, Yu_2015_CVPR}. Many have shown since that joint actor and action inference is beneficial over their independent segmentation, \eg \cite{kalogeiton2017joint,  xu2016actor}. Where all existing works learn  to segment from a fixed set of predefined actor and action pairs, we propose to segment actors and their actions in video from a natural language sentence input, as illustrated in Figure \ref{fig:intro}. 

We are inspired by recent progress in vision and language solutions for challenges like object retrieval~\cite{hu2016segmentation, hu2016natural, mao2016generation}, person search~\cite{LiPersonCVPR17, YamaguchiICCV17, ZhouAAAI17}, and object tracking~\cite{li2017tracking}. To arrive at object segmentation from a sentence, Hu \etal \cite{hu2016segmentation} rely on an LSTM network to encode an input sentence into a vector representation, before a fully convolutional network extracts a spatial feature map from an image and outputs an upsampled response map for the target object. Li \etal \cite{li2017tracking} propose object tracking from a sentence. Without specifying a bounding box, they identify a target object from the sentence and track it throughout a video. The target localization of their network is similar to Hu \etal \cite{hu2016segmentation}, be it that they introduce a dynamic convolutional layer to allow for dynamic adaptation of visual filters based on the input sentence. In effect making the textual embedding convolutional before the matching. Like \cite{hu2016segmentation,li2017tracking} we also propose an end-to-end trainable solution for segmentation from a sentence that embeds text and images into a joint model. Rather than relying on LSTMs we prefer a fully-convolutional model from the start, including dynamic filters. Moreover, we optimize our model for the task of segmenting an actor and its action in video, rather than in an image, allowing us to exploit both RGB and Flow. 

The first and foremost contribution of this paper is the new task of actor and action segmentation from a sentence. As a second contribution we propose a fully-convolutional model for pixel-level actor and action segmentation using an encoder-decoder neural architecture that is optimized for video and end-to-end trainable. Third, to show the potential of actor and action segmentation from a sentence we extend the A2D~\cite{xu2015fly} and J-HMDB~\cite{JhuangICCV2013} datasets with more than 7,500 textual sentences describing the actors and actions appearing in the video content. And finally, our experiments demonstrate the quality of the sentence-guided segmentations, the generalization ability of our model, and its advantage for traditional actor and action segmentation compared to the state-of-the-art. Before detailing our model, we first discuss related work.

\section{Related Work}\label{sec:related}

\subsection{Actor and action segmentation}
Xu \etal \cite{xu2015fly} pose the problem of actor and action segmentation in video and introduce the challenging Actor-Action Dataset (A2D) containing a fixed vocabulary of 43 actor and action pairs. They build a multi-layer conditional random field model and assign to each supervoxel from a video a label from an actor-action product space. In~\cite{xu2016actor}, Xu and Corso propose a grouping process to add long-ranging interactions to the conditional random field. Yan \etal \cite{YanCVPR17} show a multi-task ranking model atop supervoxel features allows for weakly-supervised actor and action segmentation using only video-level tags for training. Rather than relying on supervoxels, Kalogeiton~\etal~\cite{kalogeiton2017joint} propose a multi-task network architecture to jointly train an actor and action detector for a video. They extend their bounding box detections to pixel-wise segmentations by using state-of-the-art segmentation proposals~\cite{pinheiro2016learning} afterwards. 

The above works are limited to model interactions between actors and actions from a fixed predefined set of label pairs. Our work models the joint actor and action space using an open set of labels as rich as language. This has the advantage that we are able to distinguish between fine-grained actors in the same super-category, \eg a parrot or a duck rolling, and identify different actor and action instances. Thanks to a pre-trained word embedding, our model is also able to infer the segmentation from words that are outside of the actor and action vocabulary but exist in the embedding. Instead of generating intermediate supervoxels or segmentation proposals for a video, we follow a pixel-level model using an encoder-decoder neural architecture that is completely end-to-end trainable.

\subsection{Actor localization from a sentence} 
Recently, works appeared that localize a human actor from an image \cite{LiPersonCVPR17} or video~\cite{YamaguchiICCV17} based on a sentence. In~\cite{LiPersonCVPR17}, Li \etal introduce a person description dataset with sentence annotations and person samples from five existing person re-identification datasets. Their accompanying neural network model captures word-image relations and estimates the affinity between a sentence and a person image. Closer to our work is~\cite{YamaguchiICCV17}, where Yamaguchi \etal propose spatio-temporal person search in video. They supplement thousands of video clips from the ActivityNet dataset \cite{caba2015activitynet} with person descriptions. Their person retrieval model first proposes candidate tubes, ranks them based on a query in a joint visual-textual embedding and then outputs a final ranking. 

Similar to \cite{LiPersonCVPR17,YamaguchiICCV17}, we also supplement existing datasets with sentence descriptions, in our case A2D \cite{xu2015fly} and J-HMDB \cite{JhuangICCV2013}, but for the purpose of actor \textit{and} action segmentation. Where \cite{YamaguchiICCV17} demonstrates the value of sentences describing human actors for action localization in video, we generalize to actions performed by any actor. Additionally, where \cite{LiPersonCVPR17,YamaguchiICCV17}, simplify their localization to a bounding box around the human actor of interest, we output a pixel-wise segmentation of both actor and action in video. 

\subsection{Action localization from a sentence} 
Both Gao \etal ~\cite{GaoICCV17} and Hendricks \etal~\cite{HendricksMomentsICCV17} consider retrieving a specific temporal interval containing actions via a sentence. In contrast, our work offers a unique opportunity to study spatio-temporal segmentation from a sentence, with a diverse set of actors and actions. 

Jain \etal \cite{jain2015objects2action} follow a zero-shot protocol and demonstrate spatio-temporal action localization is feasible from just a sentence describing a (previously unknown) action class. They first generate a set of action tubes, encode each of them by thousands of object classifier responses, and compute a word2vec similarity between the high-scoring object categories inside an action proposal and the action query. Mettes and Snoek~\cite{MettesICCV17} also follow a zero-shot regime and match sentences to actions in a word2vec space, but rather than relying on action proposals and object classifiers, they prefer object detectors only, allowing to query for spatio-temporal relations between human actors and objects. Different from their zero-shot setting, we operate in a supervised regime. We also aim for spatio-temporal localization of actions in video, but rather than generating bounding boxes, we prefer a pixel-wise segmentation over actions performed by any actor.

\begin{figure*}[t!]
    \centering
    \includegraphics[width=0.9\linewidth]{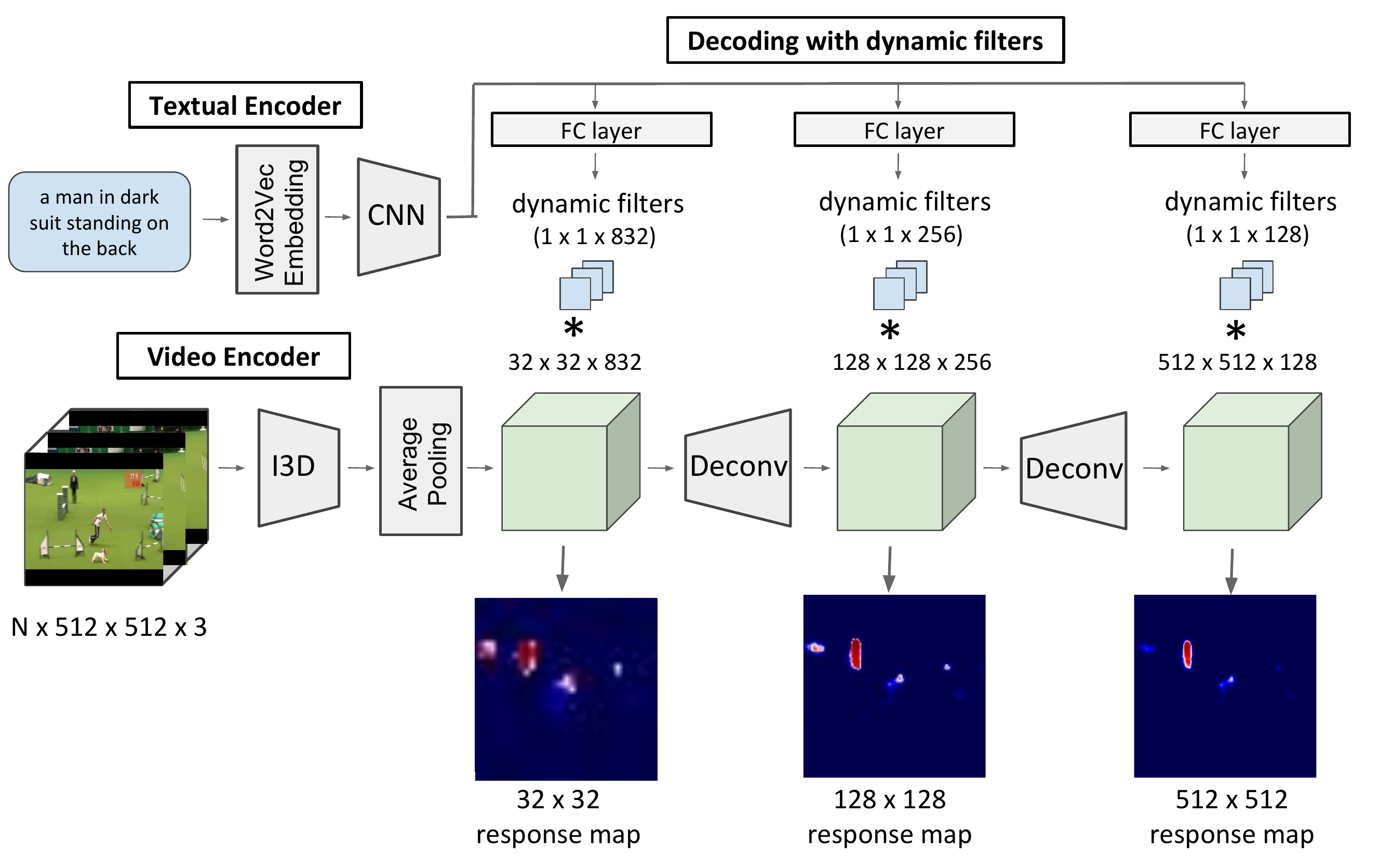}
    \caption{Our RGB model for actor and action video segmentation from a natural language sentence consists of three main components: a convolutional neural network to encode the expression, a 3D convolutional neural network to encode the video, and a decoder that performs a pixel-wise segmentation by convolving dynamic filters generated from the encoded textual representation with the encoded video representation. The same model is applied to the Flow input.}
    \label{fig:model}
\end{figure*}

\section{Model}\label{sec:proposed}
Given a video and a natural language sentence as a query, we aim to segment the actor and its action in each frame of the video as specified by the query. To achieve this, we propose a model which combines both video and language information to perform pixel-wise segmentation according to the input query. We do so by generating convolutional dynamic filters from the textual representation and convolving them with the visual representation of different resolutions to output a segmentation mask. Our model consists of three main components: a textual encoder, a video encoder and a decoder, as illustrated in Figure~\ref{fig:model}. 

\subsection{Textual Encoder}
\label{sec:proposed:textual_encoder}
Given an input natural language sentence as a query that describes the actor and action, we aim to encode it in a way that enables us to perform segmentation of the specified actor and action in video. Different from~\cite{hu2016segmentation,li2017tracking} who aim to train word embeddings from scratch on the ReferIt Dataset~\cite{sahar2014referit}, we rely on word embeddings obtained from a large collection of text documents. Particularly, we are using a word2vec model pre-trained on the Google News Dataset~\cite{mikolov2013distributed}. It enables us to handle words beyond the ones of the sentences in the training set. In addition, we are using a simple 1D convolutional neural network instead of an LSTM to encode input sentences, which we will further detail in our ablation study.

\textbf{Details.} Each word of the input sentence is represented as a 300-dimensional word2vec embedding, without any further preprocessing. All the word embeddings are fixed without fine-tuning during training. The input sentence is then represented as a concatenation of its individual word representations, \eg a 10-word sentence is represented by a $10 \times 300$ matrix. Each sentence is additionally padded to have the same size. The network consists of a single 1D convolutional layer with a temporal filter size equal to 2 and with the same output dimension as the word2vec representation. After the convolutional layer we apply the ReLU activation function and perform max-pooling to obtain a representation for the whole sentence. 

\subsection{Video Encoder}
Given an input video, we aim to obtain a visual representation that encodes both the actor and action information, while preserving the spatial information that is necessary to perform pixel-wise segmentation. Different from~\cite{hu2016segmentation,li2017tracking} who use a 2D image-based model our model takes advantage of the temporal dynamics of the video as well. Recently, Carreira and Zisserman~\cite{CarreiraCVPR2017} proposed to inflate the 2D filters of a convolutional neural network to 3D filters (I3D) to better exploit the spatio-temporal nature of video. By pre-training on both image object dataset ImageNet~\cite{russakovsky2015imagenet} and video action dataset Kinetics~\cite{kay2017kinetics} their model achieves state-of-the-art results for action classification. We adopt the I3D model to obtain a visual representation from video. 

Moreover, we also follow the well-known two-stream approach~\cite{SimonyanNIPS2014} to combine appearance and motion information, which was successfully applied earlier to a wide range of video understanding tasks such as action classification~\cite{FeichtenhoferCVPR2017, WangECCV16} and detection~\cite{PengECCV2016, ZhaoICCV2017}. We study the effect of having RGB and Flow inputs for actor and action segmentation in our ablation study.  

\textbf{Details.} Frames of all videos are padded to have the same size. As visual feature representation for both the RGB and Flow input, we use the output of the inception block before the last max-pooling layer of the I3D network followed by an average pooling over the temporal dimension. To obtain a more robust descriptor at each spatial location, $L2$-normalization is applied to every spatial position in the feature map. Following~\cite{hu2016segmentation,li2017tracking}, we also append the spatial coordinates of each position as extra channels to the visual representation to allow learning spatial qualifiers like ``left of'' or ``above''.

\subsection{Decoding with dynamic filters}
To perform pixel-wise segmentation from a natural language sentence we rely on dynamic convolutional filters, as earlier proposed in~\cite{li2017tracking}. Unlike static convolutional filters that are used in conventional convolutional neural networks, dynamic filters are generated depending on the input, in our case on the encoded sentence representation. It enables us to transfer textual information to the visual domain. Different from~\cite{li2017tracking}, we notice better results with a $\tanh$ activation function and $L2$-normalization on the features. In addition, we generate dynamic filters for several resolutions with different network parameters.

Given a sentence representation $T$, we generate dynamic filters $f^r$ for each resolution ${r\in R}$ with a separate single layer fully-connected network:
\begin{align}
f^r = \tanh(W^r_fT + b^r_f), 
\end{align}
where $\tanh$ is the hyperbolic tangent function and $f^r$ has the same number of channels as representation $V^r_t$ for video input at timestep $t$ and resolution $r$. Then the dynamic filters are convolved with $V^r_t$ to obtain a pixel-wise segmentation response map for resolution $r$ at timestep $t$: 
\begin{align}
S^r_t = f^r * V^r_t, 
\end{align}
To obtain a segmentation mask with the same resolution as the input video, we further employ a deconvolutional neural network. Different from~\cite{hu2016segmentation, li2017tracking}, who apply deconvolution on the segmentation response maps, we use the deconvolutional layers on the video representation $V^r_t$ directly.
It enables us to better handle small objects and output smoother segmentation predictions. In addition, it helps to obtain more accurate segmentations for high overlap values as we will show in the experiments. 

\textbf{Details.} Each of our deconvolutional networks consists of two blocks with one deconvolutional layer with kernel size $8 \times 8$ and stride $4$, followed by a convolutional layer with a kernel size of $3 \times 3$ and a stride of $1$. We use only the highest-resolution response map for the final segmentation prediction.

\subsection{Training}
\label{sec:proposed:training}
Our training sample consists of an input video clip, an input sentence and a binary ground truth segmentation mask $Y^r$ for each resolution $r \in R$ of the frame in the middle of each input video clip. For each training sample we define a loss, while taking into account multiple resolutions, which helps for better flow of gradients in the model similar to a skip-connection approach: 
\begin{align}
\mathcal{L} = \sum_{r\in R}{\alpha_{r}\mathcal{L}^r} 
\end{align}
\begin{align}
\mathcal{L}^r = \frac{1}{r^2}\sum_{i=1}^{r}\sum_{j=1}^{r}{\mathcal{L}_{ij}^{r}}
\end{align}
where $\alpha_r$ is a weight for resolution $r$. In this paper we consider $R = \{32, 128, 512\}$ and we further discuss the importance of using losses of all resolutions in our ablation study. 

The pixel-wise $\mathcal{L}_{ij}^{r}$ loss is a logistic loss defined as follows:
\begin{align}
\mathcal{L}_{ij}^r = log(1 + \exp{(-S_{ij}^rY_{ij}^r)})
\end{align}
where $S_{ij}^r$ is a response value of our model at pixel $(i, j)$ for resolution $r$ and $Y_{ij}^r$ is a binary label at pixel $(i, j)$ for resolution $r$.  

\textbf{Details.} We train our model using the Adam optimizer~\cite{KingmaICLR15} with a learning rate of $0.001$ and other parameters of the optimizer set to the default values. We divide the learning rate by $10$ every $5,000$ iterations and train for $15,000$ iterations in total. We finetune only the last inception block of the video encoder.

\section{Datasets}\label{sec:experiments:datasets}

\subsection{A2D Sentences}
The Actor-Action Dataset (A2D) by Xu \etal~\cite{xu2015fly} serves as the largest video dataset for the general actor and action segmentation task. It contains 3,782 videos from YouTube with pixel-level labeled actors and their actions. The dataset includes eight different actions, while a total of seven actor classes are considered to perform those actions. We follow~\cite{xu2015fly}, who split the dataset into 3,036 training videos and 746 testing videos.

As we are interested in pixel-level actor and action segmentation from sentences, we augment the videos in A2D with natural language descriptions about what each actor is doing in the videos. Following the guidelines set forth in~\cite{sahar2014referit}, we ask our annotators for a discriminative referring expression of each actor instance if multiple objects are considered in a video. The annotation process resulted in a total of 6,656 sentences, including 811 different nouns, 225 verbs and 189 adjectives. Our sentences enrich the actor and action pairs from the A2D dataset with finer granularities. For example, the actor \textit{adult} in A2D may be annotated with \textit{man}, \textit{woman}, \textit{person} and \textit{player} in our sentences, while action \textit{rolling} may also refer to \textit{flipping}, \textit{sliding}, \textit{moving} and \textit{running} when describing different actors in different scenarios. Our sentences contain on average more words than the ReferIt dataset~\cite{sahar2014referit} (7.3 vs 4.7), even when we leave out prepositions, articles and linking verbs (4.5 vs 3.6). This makes sense as our sentences contain a variety of verbs while existing referring expression datasets mostly ignore verbs.

\subsection{J-HMDB Sentences} 
J-HMDB~\cite{JhuangICCV2013} contains 928 video clips of 21 different actions annotated with a 2D articulated human puppet that provides scale, pose, segmentation and a coarse viewpoint for the humans involved in each action. We augment the videos with sentences following the same protocol as for A2D Sentences. We ask annotators to return a natural language description of what the target object is doing in each video. We obtain 928 sentences, including 158 different nouns, 53 verbs and 23 adjectives. The most popular actors are \textit{man}, \textit{woman}, \textit{boy}, \textit{girl} and \textit{player}, while \textit{shooting}, \textit{pouring}, \textit{playing}, \textit{catching} and \textit{sitting} are the most popular actions. 

We show sentence-annotated examples of both datasets in Figure \ref{fig:datasets} and provide more details on the datasets in the supplemental material. 
The sentence annotations and the code of our model will be available at \url{https://kgavrilyuk.github.io/publication/actor_action/}.

\begin{figure}[t!]
    \centering 
    \includegraphics[width=1.0\linewidth]{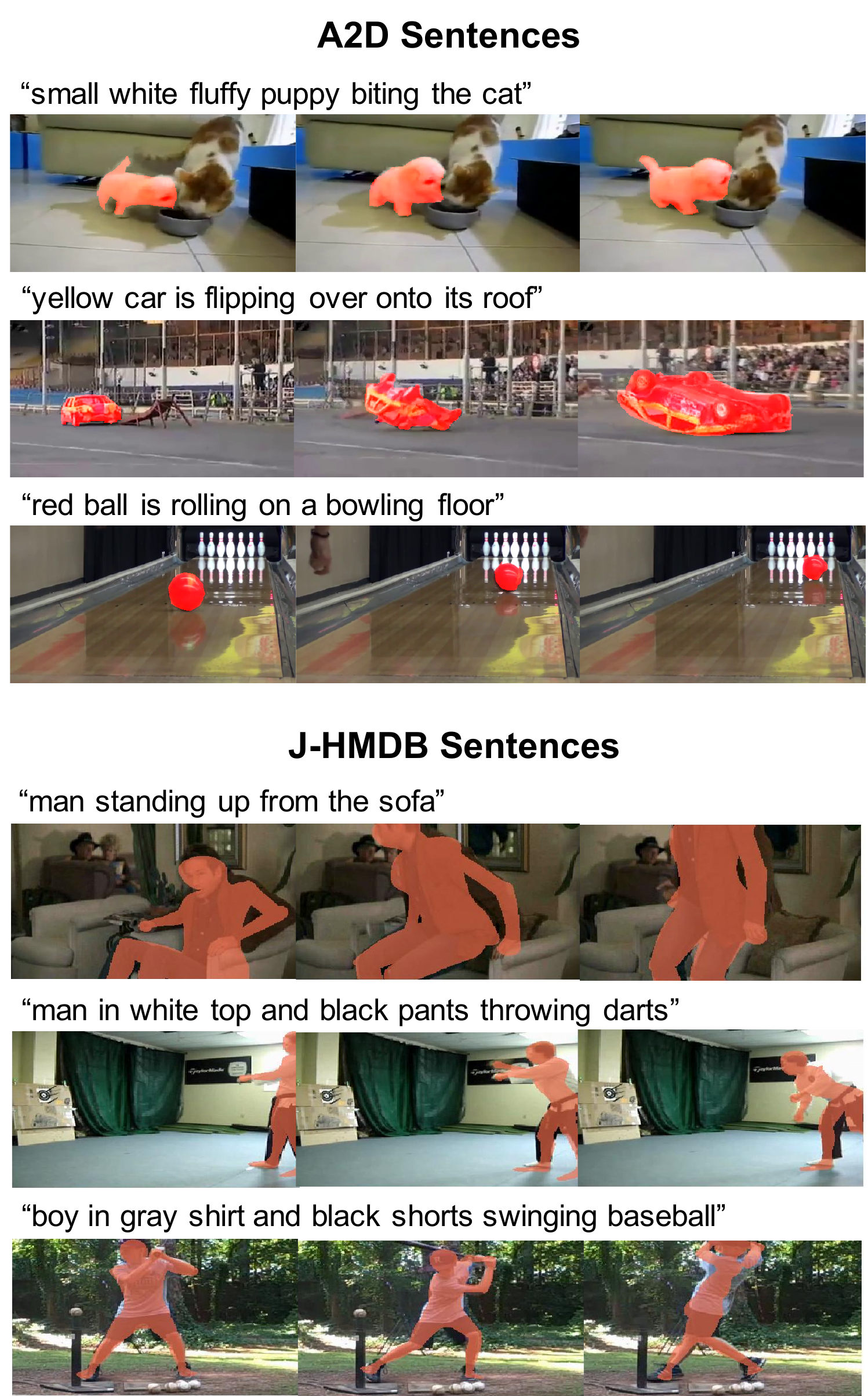} 
    \caption{A2D Sentences and J-HMDB Sentences example videos, ground truth segments and sentence annotations.}
    \label{fig:datasets}
\end{figure}

\section{Experiments}\label{sec:experiments}

\begin{table*}[t!]
\renewcommand{\arraystretch}{1.1}
\centering
\begin{tabular}{l*{9}{r}}
\toprule
&  \multicolumn{5}{c}{\textbf{Overlap}} & \multicolumn{1}{c}{\textbf{mAP}} & \multicolumn{2}{c} {\textbf{IoU}} \\
			& P@0.5 & P@0.6 & P@0.7 & P@0.8 & P@0.9 & 0.5:0.95 & Overall & Mean \\
\cmidrule(lr){2-6} \cmidrule(lr){7-7} \cmidrule(lr){8-9}
Hu~\etal~\cite{hu2016segmentation} & 7.7 & 3.9 & 0.8 & 0.0 & 0.0 & 2.0 & 21.3 & 12.8 \\
Li~\etal~\cite{li2017tracking} & 10.8 & 6.2 & 2.0 & 0.3 & 0.0 & 3.3 & 24.8 & 14.4  \\
Hu~\etal~\cite{hu2016segmentation}~$\star$ & 34.8 & 23.6 & 13.3 & 3.3 & 0.1 & 13.2 & 47.4 & 35.0 \\
Li~\etal~\cite{li2017tracking}~$\star$ & 38.7 & 29.0 & 17.5 & 6.6 & 0.1 & 16.3 & 51.5 & 35.4 \\
\cmidrule{1-9}
\textit{This paper: RGB} & 47.5 & 34.7 & 21.1 & 8.0 & 0.2 & 19.8 & 53.6 & 42.1 \\ 
\textit{This paper: RGB + Flow} & \textbf{50.0} & \textbf{37.6} & \textbf{23.1} & \textbf{9.4} & \textbf{0.4} & \textbf{21.5} & \textbf{55.1} & \textbf{42.6} \\
\bottomrule
\end{tabular}
\smallskip
\caption{Segmentation from a sentence on A2D Sentences. Object segmentation baselines~\cite{hu2016segmentation,li2017tracking} as proposed in the original papers, or fine-tuned on the A2D Sentences train split (denoted by $\star$). Our model outperforms both baselines for all metrics. Incorporating Flow in our video model further improves results.}
\label{table:experiments:actoraction_res}
\end{table*}

\subsection{Ablation Study} 

In the first set of experiments we study the impact of individual components on our proposed model.

\textbf{Setup.} We select A2D Sentences for these set of experiments and use the train split for training and the test split for evaluation. The input to our model is a sentence describing what to segment and a video clip of $N$ RGB frames around the frame to be segmented.

\textbf{Evaluation.} We adopt the widely used intersection-over-union (IoU) metric to measure segmentation quality. As aggregation metric we consider \textit{overall IoU}, which is computed as total intersection area of all test data over the total union area.

\textbf{Results on A2D Sentences.} We first evaluate the influence of the number of input frames on our visual encoder and the segmentation result. We run our model with $N = 1, 4, 8, 16$ and we get $48.2\%$, $52.2\%$, $52.8\%$, and $53.6\%$ respectively in terms of \textit{overall IoU}. It reveals the important role of the large temporal context for actor and action video segmentation. Therefore, we choose $N = 16$ for all remaining experiments.

Next we compare our 1D convolutional textual encoder with an LSTM encoder. We follow the same setting for LSTM as in~\cite{hu2016segmentation,li2017tracking}, we use a final hidden state of LSTM as textual representation for the whole sentence. The dimension of the hidden state is set to $1,000$. We represent words by the same word2vec embedding model for both models. We observe that our simple 1D convolutional textual encoder outperforms LSTM in terms of \textit{overall IoU}: $53.6\%$ for our encoder and $51.8\%$ for LSTM. We also experimented with bidirectional LSTM which slightly improves results over vanilla LSTM to $52.1\%$. Therefore, we select the convolutional neural network to encode the textual input in the remaining experiments.

We further investigate the importance of our multi-resolution loss. We compare the setting when we are using all three resolutions to compute the loss ($\alpha_r = 1, r \in \{32, 128, 512\}$) with the setting when only the highest resolution is used ($\alpha_{32, 128} = 0, \alpha_{512} = 1$). In terms of \textit{overall IoU} the multi-resolution setting performs $53.6\%$ while single resolution performs $49.4\%$. This demonstrates the benefit of the multi-resolution loss in our model.

\begin{figure*}[t!]
    \centering 
    \includegraphics[width=0.99\textwidth]{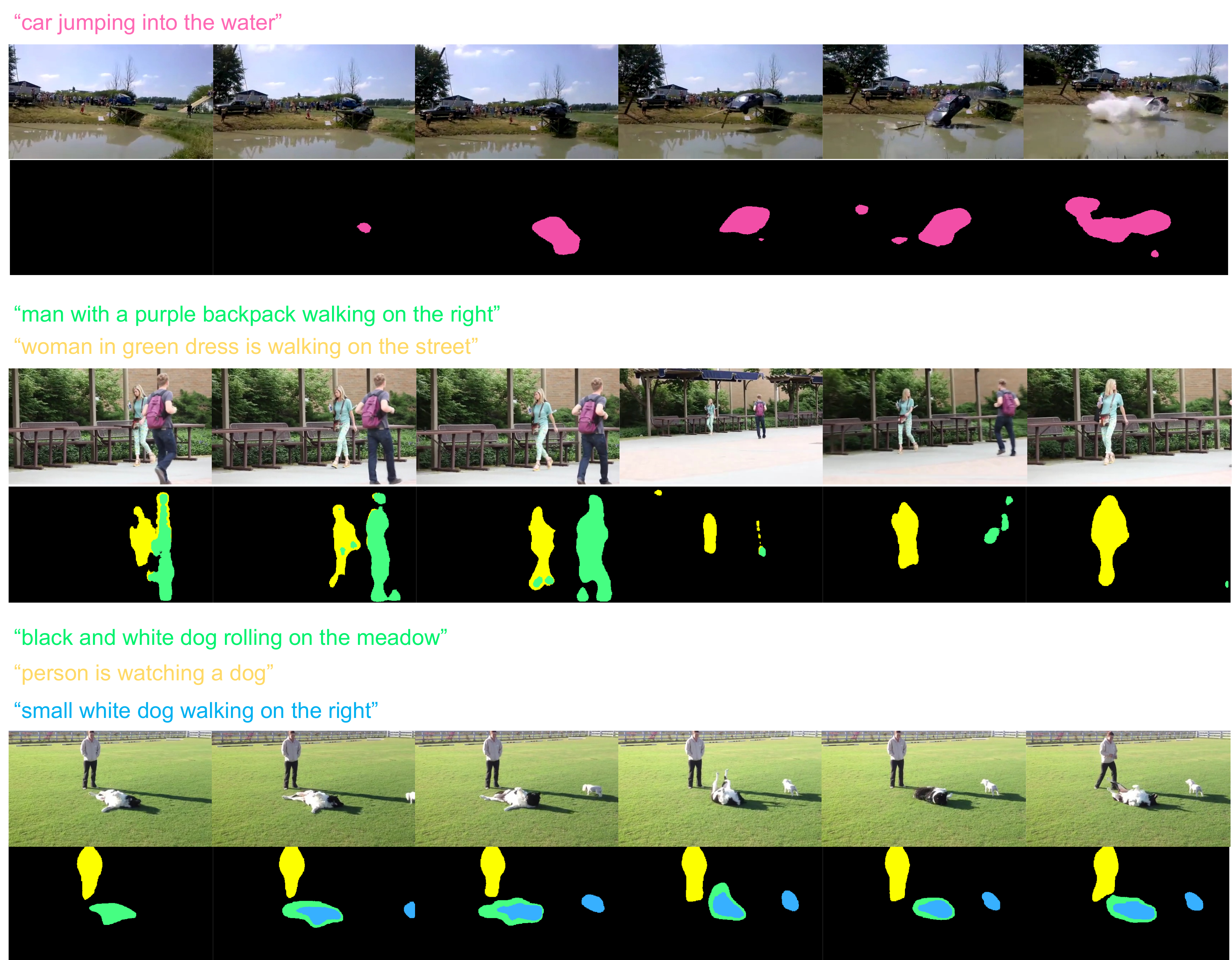} 
    \caption{Visualized segmentation results from our model on A2D Sentences. The first row shows a video with single actor and action, while the video in the second row contains similar types of actors performing the same action. In the third row, we illustrate a video with three sentences describing not only different actors, but also the same type of actor performing different actions. The colored segmentation masks are generated from the sentence with the same color above each video.}
    \label{fig:visualization}
\end{figure*}

In the last experiment we study the impact of the two-stream~\cite{SimonyanNIPS2014} approach for our task. We make a comparison for two type of inputs - RGB and Flow. For both streams we use 16 frames as input. The RGB stream produces better results than Flow: $53.6\%$ for RGB and $49.5\%$ for Flow. We then explore a fusion of RGB and Flow streams by computing a weighted average of the response maps from each stream. When we set the weight for RGB 2 times larger than Flow, it further improves our results to $55.1\%$.

\subsection{Segmentation from a sentence}
\label{sec:experiments:actoraction}
In this experiment, we segment a video based on a given natural language sentence on the newly annotated A2D Sentences and J-HMDB Sentences datasets and compare our proposed model with the baseline methods.

\textbf{Setup.} As there is no prior work for video segmentation from a sentence, we select two methods~\cite{hu2016segmentation, li2017tracking}, which can be used for the related task of image segmentation from a sentence, as our baselines. To be precise, we compare with the segmentation model of~\cite{hu2016segmentation} and the lingual specification model of~\cite{li2017tracking}. We report baseline results in two training settings. In the first one, the baselines are trained solely on the ReferIt dataset~\cite{sahar2014referit}, as indicated in the original papers. In the second setting we further fine-tune the baseline models using the training videos from A2D Sentences. We train our model only on the train split of A2D Sentences. During test, we follow~\cite{xu2015fly} and evaluate the models on each frame of the test videos for which segmentation annotation is available - around one to three frames per video. The input to both baseline models is an RGB frame with a sentence description. For our model, we use the same sentence as input but instead of a single RGB frame we employ 16 frames around the frame to be segmented as this setting shows the best results in our ablation study.

\textbf{Evaluation.} In addition to \textit{overall IoU}, we also consider \textit{mean IoU} as aggregation. The \textit{mean IoU} is computed as the average over the IoU of each test sample.
While the \textit{overall IoU} favors large segmented regions, \textit{mean IoU} treats large and small regions equally. In addition, following~\cite{hu2016segmentation,li2017tracking}, we also measure precision at five different overlap values ranging from $0.5$ to $0.9$ as well as the mean average precision over $.50:.05:.95$~\cite{lin2014microsoft}.

\textbf{Results on A2D Sentences.} In Table~\ref{table:experiments:actoraction_res}, we report the results on the A2D Sentences dataset. 
The model of~\cite{hu2016segmentation} and~\cite{li2017tracking}, pretrained on ReferIt~\cite{sahar2014referit}, performs modestly as this dataset contains rich sentences describing objects, but it provides less information about actions. Fine-tuning these two baselines on A2D Sentences helps improve their performance by incorporating the notion of actions into the models. Our model outperforms both baselines for all metrics using RGB frames as input, bringing $3.5\%$ absolute improvement in $mAP$, $2.1\%$ in \textit{overall IoU} and $6.7\%$ in \textit{mean IoU}. Fusion of RGB and Flow streams further improves our results. The larger improvement in \textit{mean IoU} compared to \textit{overall IoU} indicates our model is especially better on segmenting small objects. The results in mAP show the benefit of our model for larger overlap values. We visualize some of the sentence-guided segmentation results in Figure~\ref{fig:visualization}. First of all, our model can tackle the scenarios when the actor is not in the frame, \eg in the second video. The model stops generating the segmentation once the man has left the camera's view.  Our model can also tackle the scenarios when the actor is performing an action which is different from the one specified in the sentence, \eg in the first video. The model doesn't output any segmentation for the frames in which the car is not in the \textit{jumping} state. It shows the potential of our model for spatio-temporal video segmentation. Second, in contrast to segmentation from actor-action labels, we can see from the second video that our segmentation from a sentence enables to distinguish the instances of the same actor-action pair by richer descriptions. In the third video, our model confuses two dogs, still we easily segment different types of actors.

\textbf{Results on J-HMDB Sentences.} We further evaluate the generalization ability of our model and the baselines. We test the models, finetuned or trained on A2D Sentences, on all $928$ videos of J-HMDB Sentences dataset without any additional finetuning. For each video, we uniformly sample three frames for evaluation following the same setting as in the previous experiment. We report our results in Table~\ref{table:experiments:jhmdb_res}.

J-HMDB Sentences focuses exclusively on human actions and $4$ out of $21$ actions overlap with actions in A2D Sentences, namely \textit{climb stairs}, \textit{jump}, \textit{walk}, and \textit{run}. 
Consistent with the results on A2D Sentences, our method provides a more accurate segmentation for higher overlap values which is shown by \textit{mAP}. We attribute the better generalization ability to two aspects. The baselines rely on the VGG16~\cite{simonyan2014very} model to represent images, while we are using the video-specific I3D model. The second aspect comes from our textual representation, which can exploit similarity in descriptions of A2D Sentences and J-HMDB Sentences.

\begin{table*}[t]
\renewcommand{\arraystretch}{1.1}
\centering
\begin{tabular}{l*{9}{r}}
\toprule
&  \multicolumn{5}{c}{\textbf{Overlap}} & \multicolumn{1}{c}{\textbf{mAP}} & \multicolumn{2}{c}{\textbf{IoU}} \\
			& P@0.5 & P@0.6 & P@0.7 & P@0.8 & P@0.9 & 0.5:0.95 & Overall & Mean\\
\cmidrule(lr){2-6} \cmidrule(lr){7-7} \cmidrule(lr){8-9} 
Hu~\etal~\cite{hu2016segmentation} & 63.3 & 35.0 & 8.5 & 0.2 & 0.0 & 17.8 & \textbf{54.6} & 52.8  \\
Li~\etal~\cite{li2017tracking} & 57.8 & 33.5 & 10.3 & 0.6 & 0.0 & 17.3 & 52.9 & 49.1 \\
\cmidrule{1-9}
\textit{This paper} & \textbf{69.9} & \textbf{46.0} & \textbf{17.3} & \textbf{1.4} &  0.0 & \textbf{23.3} & 54.1 & \textbf{54.2} \\
\bottomrule
\end{tabular}
\smallskip
\caption{Segmentation from a sentence on J-HMDB Sentences using best settings per model on A2D Sentences, demonstrating generalization ability. Our model generates more accurate segmentations for higher overlap values.}
\label{table:experiments:jhmdb_res}
\end{table*}

\begin{table*}[t!]
\renewcommand{\arraystretch}{1.1}
\centering
\scalebox{0.84}{
\begin{tabular}{l*{10}{r}}
\toprule
&  \multicolumn{3}{c}{\textbf{Actor}} & \multicolumn{3}{c}{\textbf{Action}} & \multicolumn{3}{c}{\textbf{Actor and Action}} \\
			& Class-Average  & Global & Mean IoU & Class-Average & Global & Mean IoU & Class-Average & Global & Mean IoU\\
\cmidrule(lr){2-4} \cmidrule(lr){5-7} \cmidrule(lr){8-10} 
Xu~\etal~\cite{xu2015fly} & 45.7 & 74.6 & - & 47.0 & 74.6 & - & 25.4 & 76.2 & -  \\
Xu~\etal~\cite{xu2016actor}  & 58.3 & 85.2 & 33.4 & 60.5 & 85.3 & 32.0 & 43.3 & 84.2 & 19.9   \\
Kalogeiton~\etal~\cite{kalogeiton2017joint} & \textbf{73.7} & 90.6 & 49.5 & 60.5 & 89.3 & 42.2 & 47.5 & 88.7 & 29.7  \\
\cmidrule{1-10}
\textit{This paper} & 71.4 & \textbf{92.8} & \textbf{53.7} & \textbf{69.3} & \textbf{92.5} & \textbf{49.4} & \textbf{52.4} & \textbf{91.7} & \textbf{34.8} \\
\bottomrule
\end{tabular}
}
\smallskip
\caption{Semantic segmentation results on the A2D dataset using actor, action and actor+action as input respectively. Even though our method is not designed for this setting, it outperforms the state-of-the-art in most of the cases.}
\label{table:experiments:semanticsegm_res}
\end{table*}

\subsection{Segmentation from actor and action pairs}
\label{sec:experiments:semantic}
Finally, we segment a video from a predefined set of actor and action pairs and compare it with the state-of-the-art segmentation models on the original A2D dataset~\cite{xu2015fly}.

\textbf{Setup.} Instead of input sentences, we train our model on the $43$ valid actor and action pairs provided by the dataset, such as \textit{adult walking} and \textit{dog rolling}. We use these pairs as textual input to our model. Visual input is kept the same as before. 
As our model explicitly requires a textual input for a given video, we select a subset of pairs from all possible pairs as queries to our model. For this purpose, we finetune a multi-label classification network on A2D dataset and select the pairs with a confidence score higher than $0.5$. We use this reduced set of pairs as queries to our model and pick the class label with the highest response for each pixel. The classification network contains an RGB and a Flow I3D model where the number of neurons in the last layer is set to $43$ and the activation function is replaced by a $sigmoid$ for multi-label classification. During training, we finetune the last inception block and the final layer of both models on random 64-frame video clips. We randomly flip each frame horizontally in the video clip and then extract a $224 \times 224$ random crop. We train for $3,000$ iterations with the Adam optimizer and fix the learning rate to $0.001$. During test, we extract 32-frame clips over the video and average the scores across all the clips and across RGB and Flow streams to obtain the final score for a given video. For this multi-label classification we obtain mean average precision of $70\%$, compared to $67\%$ in~\cite{xu2015fly}.

\textbf{Evaluation.} We report the class-average pixel accuracy, global pixel accuracy and \textit{mean IoU} as in~\cite{kalogeiton2017joint}. Pixel accuracy is the percentage of pixels for which the label is correctly predicted, either over all pixels (global) or first computed for each class separately and then averaged over classes (class-average).

\textbf{Results on A2D.} We compare our approach with the state-of-the-art in Table~\ref{table:experiments:semanticsegm_res}. Even though our method is not designed for this setting, it outperforms all the competitors for joint actor and action segmentation (last $3$ columns of Table~\ref{table:experiments:semanticsegm_res}). Particularly, we improve the state-of-the-art by a margin of $4.9\%$ in terms of class-average accuracy and $5.1\%$ in terms of Mean IoU.
In addition to joint actor and action segmentation, we report results for actor and action segmentation separately. For actor segmentation the method by Kalogeiton~\etal~\cite{kalogeiton2017joint} is slightly better in terms of class-average accuracy, for all other metrics and settings our method sets a new state-of-the-art. Our improvement is particularly notable on action segmentation where we outperform the state-of-the-art by $8.8\%$ in terms of class-average accuracy and $7.2\%$ in terms of Mean IoU. It validates that our method is suitable for both actor and action segmentation, be it individually or combined.

\section{Conclusion}\label{sec:conclusion}
We introduce the new task of actor and action video segmentation from a sentence. Our encoder-decoder neural architecture for pixel-level segmentation explicitly takes into account the spatio-temporal nature of video. To enable sentence-guided segmentation with our model, we extended two existing datasets with sentence-level annotations describing actors and their actions in the video content. Experiments show the feasibility and robustness, as well as the model's ability to adapt to the task of semantic segmentation of actor and action pairs, outperforming the state-of-the-art.

{\small
\bibliographystyle{ieee}
\bibliography{egbib}

\begin{thebibliography}{1}\itemsep=-1pt

\bibitem{supp_hu2016segmentation}
R.~Hu, M.~Rohrbach, and T.~Darrell.
\newblock Segmentation from natural language expressions.
\newblock In {\em ECCV}, 2016.

\bibitem{supp_li2017tracking}
Z.~Li, R.~Tao, E.~Gavves, C.~G.~M. Snoek, and A.~W.~M. Smeulders.
\newblock Tracking by natural language specification.
\newblock In {\em CVPR}, 2017.

\bibitem{supp_xu2015fly}
C.~Xu, S.-H. Hsieh, C.~Xiong, and J.~J. Corso.
\newblock Can humans fly? action understanding with multiple classes of actors.
\newblock In {\em CVPR}, 2015.

\end{thebibliography}


\begin{thebibliography}{10}\itemsep=-1pt

\bibitem{caba2015activitynet}
F.~Caba~Heilbron, V.~Escorcia, B.~Ghanem, and J.~C. Niebles.
\newblock Activitynet: A large-scale video benchmark for human activity
  understanding.
\newblock In {\em CVPR}, 2015.

\bibitem{CarreiraCVPR2017}
J.~Carreira and A.~Zisserman.
\newblock Quo vadis, action recognition? a new model and the kinetics dataset.
\newblock In {\em CVPR}, 2017.

\bibitem{FeichtenhoferCVPR2017}
C.~Feichtenhofer, A.~Pinz, and R.~P. Wildes.
\newblock Spatiotemporal multiplier networks for video action recognition.
\newblock In {\em CVPR}, 2017.

\bibitem{GaoICCV17}
J.~Gao, C.~Sun, Z.~Yang, and R.~Nevatia.
\newblock Tall: Temporal activity localization via language query.
\newblock In {\em ICCV}, 2017.

\bibitem{HendricksMomentsICCV17}
L.~A. Hendricks, O.~Wang, E.~Shechtman, J.~Sivic, T.~Darrell, and B.~Russell.
\newblock Localizing moments in video with natural language.
\newblock In {\em ICCV}, 2017.

\bibitem{hu2016segmentation}
R.~Hu, M.~Rohrbach, and T.~Darrell.
\newblock Segmentation from natural language expressions.
\newblock In {\em ECCV}, 2016.

\bibitem{hu2016natural}
R.~Hu, H.~Xu, M.~Rohrbach, J.~Feng, K.~Saenko, and T.~Darrell.
\newblock Natural language object retrieval.
\newblock In {\em CVPR}, 2016.

\bibitem{jain2015objects2action}
M.~Jain, J.~van Gemert, T.~Mensink, and C.~G.~M. Snoek.
\newblock Objects2action: Classifying and localizing actions without any video
  example.
\newblock In {\em ICCV}, 2015.

\bibitem{JhuangICCV2013}
H.~Jhuang, J.~Gall, S.~Zuffi, C.~Schmid, and M.~J. Black.
\newblock Towards understanding action recognition.
\newblock In {\em ICCV}, 2013.

\bibitem{kalogeiton2017joint}
V.~Kalogeiton, P.~Weinzaepfel, V.~Ferrari, and C.~Schmid.
\newblock Joint learning of object and action detectors.
\newblock In {\em ICCV}, 2017.

\bibitem{kay2017kinetics}
W.~Kay, J.~Carreira, K.~Simonyan, B.~Zhang, C.~Hillier, S.~Vijayanarasimhan,
  F.~Viola, T.~Green, T.~Back, P.~Natsev, M.~Suleyman, and A.~Zisserman.
\newblock The kinetics human action video dataset.
\newblock {\em arXiv preprint arXiv:1705.06950}, 2017.

\bibitem{sahar2014referit}
S.~Kazemzadeh, V.~Ordonez, M.~Matten, and T.~L. Berg.
\newblock {ReferIt} game: Referring to objects in photographs of natural
  scenes.
\newblock In {\em EMNLP}, 2014.

\bibitem{KingmaICLR15}
D.~P. Kingma and J.~Ba.
\newblock Adam: A method for stochastic optimization.
\newblock In {\em ICLR}, 2015.

\bibitem{LiPersonCVPR17}
S.~Li, T.~Xiao, H.~Li, B.~Zhou, D.~Yue, and X.~Wang.
\newblock Person search with natural language description.
\newblock In {\em CVPR}, 2017.

\bibitem{li2017tracking}
Z.~Li, R.~Tao, E.~Gavves, C.~G.~M. Snoek, and A.~W.~M. Smeulders.
\newblock Tracking by natural language specification.
\newblock In {\em CVPR}, 2017.

\bibitem{lin2014microsoft}
T.-Y. Lin, M.~Maire, S.~Belongie, J.~Hays, P.~Perona, D.~Ramanan,
  P.~Doll{\'a}r, and C.~L. Zitnick.
\newblock Microsoft coco: Common objects in context.
\newblock In {\em ECCV}, 2014.

\bibitem{mao2016generation}
J.~Mao, H.~Jonathan, A.~Toshev, O.~Camburu, A.~Yuille, and K.~Murphy.
\newblock Generation and comprehension of unambiguous object descriptions.
\newblock In {\em CVPR}, 2016.

\bibitem{MettesICCV17}
P.~Mettes and C.~G.~M. Snoek.
\newblock Spatial-aware object embeddings for zero-shot localization and
  classification of actions.
\newblock In {\em ICCV}, 2017.

\bibitem{MettesBMVC17}
P.~Mettes, C.~G.~M. Snoek, and S.-F. Chang.
\newblock Localizing actions from video labels and pseudo-annotations.
\newblock In {\em BMVC}, 2017.

\bibitem{mikolov2013distributed}
T.~Mikolov, I.~Sutskever, K.~Chen, G.~S. Corrado, and J.~Dean.
\newblock Distributed representations of words and phrases and their
  compositionality.
\newblock In {\em NIPS}, 2013.

\bibitem{PengECCV2016}
X.~Peng and C.~Schmid.
\newblock Multi-region two-stream r-cnn for action detection.
\newblock In {\em ECCV}, 2016.

\bibitem{pinheiro2016learning}
P.~O. Pinheiro, T.-Y. Lin, R.~Collobert, and P.~Doll{\'a}r.
\newblock Learning to refine object segments.
\newblock In {\em ECCV}, 2016.

\bibitem{russakovsky2015imagenet}
O.~Russakovsky, J.~Deng, H.~Su, J.~Krause, S.~Satheesh, S.~Ma, Z.~Huang,
  A.~Karpathy, A.~Khosla, M.~Bernstein, A.~Berg, and L.~Fei-Fei.
\newblock {ImageNet} large scale visual recognition challenge.
\newblock {\em IJCV}, 2015.

\bibitem{SimonyanNIPS2014}
K.~Simonyan and A.~Zisserman.
\newblock Two-stream convolutional networks for action recognition in videos.
\newblock In {\em NIPS}, 2014.

\bibitem{simonyan2014very}
K.~Simonyan and A.~Zisserman.
\newblock Very deep convolutional networks for large-scale image recognition.
\newblock In {\em ICLR}, 2015.

\bibitem{TianCVPR13}
Y.~Tian, R.~Sukthankar, and M.~Shah.
\newblock Spatiotemporal deformable part models for action detection.
\newblock In {\em CVPR}, 2013.

\bibitem{WangECCV16}
L.~Wang, Y.~Xiong, Z.~Wang, Y.~Qiao, D.~Lin, X.~Tang, and L.~{Van Gool}.
\newblock Temporal segment networks: Towards good practices for deep action
  recognition.
\newblock In {\em ECCV}, 2016.

\bibitem{xu2016actor}
C.~Xu and J.~J. Corso.
\newblock Actor-action semantic segmentation with grouping process models.
\newblock In {\em CVPR}, 2016.

\bibitem{xu2015fly}
C.~Xu, S.-H. Hsieh, C.~Xiong, and J.~J. Corso.
\newblock Can humans fly? action understanding with multiple classes of actors.
\newblock In {\em CVPR}, 2015.

\bibitem{YamaguchiICCV17}
M.~Yamaguchi, K.~Saito, Y.~Ushiku, and T.~Harada.
\newblock Spatio-temporal person retrieval via natural language queries.
\newblock In {\em ICCV}, 2017.

\bibitem{YanCVPR17}
Y.~Yan, C.~Xu, D.~Cai, and J.~Corso.
\newblock Weakly supervised actor-action segmentation via robust multi-task
  ranking.
\newblock In {\em CVPR}, 2017.

\bibitem{Yu_2015_CVPR}
G.~Yu and J.~Yuan.
\newblock Fast action proposals for human action detection and search.
\newblock In {\em CVPR}, 2015.

\bibitem{ZhaoICCV2017}
Y.~Zhao, Y.~Xiong, L.~Wang, Z.~Wu, X.~Tang, and D.~Lin.
\newblock Temporal action detection with structured segment networks.
\newblock In {\em ICCV}, 2017.

\bibitem{ZhouAAAI17}
T.~Zhou and J.~Yu.
\newblock Natural language person retrieval.
\newblock In {\em AAAI}, 2017.

\end{thebibliography}
}

\newpage

\begin{center}
\textbf{\large Supplementary material for: \\ Actor and Action Video Segmentation from a Sentence}
\end{center}
\setcounter{equation}{0}
\setcounter{figure}{0}
\setcounter{table}{0}
\setcounter{section}{0}
\setcounter{page}{1}
\makeatletter
\renewcommand{\theequation}{S\arabic{equation}}
\renewcommand{\thefigure}{S\arabic{figure}}
\renewcommand{\thesection}{S\arabic{section}}

In this supplementary material, we first report annotation statistics on both the A2D Sentences and J-HMDB Sentences datasets in Section~\ref{sec:datasets}. In Section~\ref{sec:res_1_a2d_sentences}, we show more segmentation results of our proposed model followed by a qualitative comparison of our video-based model with the image-based models of Hu~\etal~\citeS{supp_hu2016segmentation} and Li~\etal~\citeS{supp_li2017tracking} in Section~\ref{sec:res_2_comp_baseline}.
\section{Dataset statistics}\label{sec:datasets}
\begin{figure*}[t]
    \centering 
    \includegraphics[width=\textwidth, height=0.22\textheight]{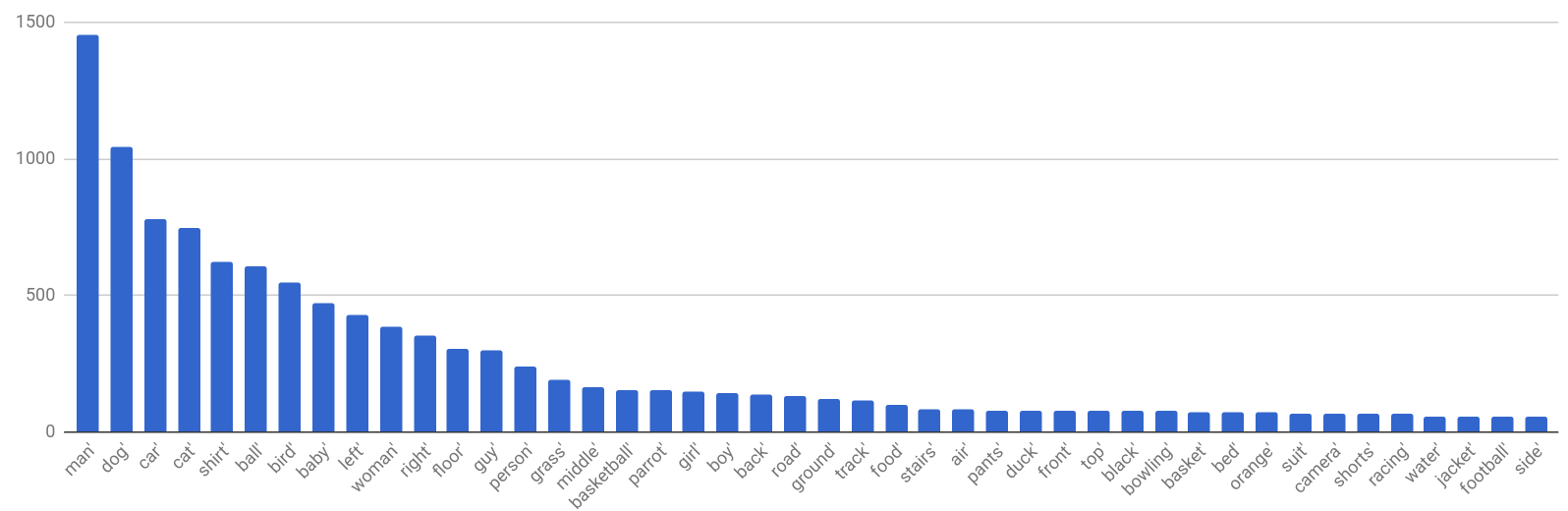}
    
    \includegraphics[width=\textwidth, height=0.22\textheight]{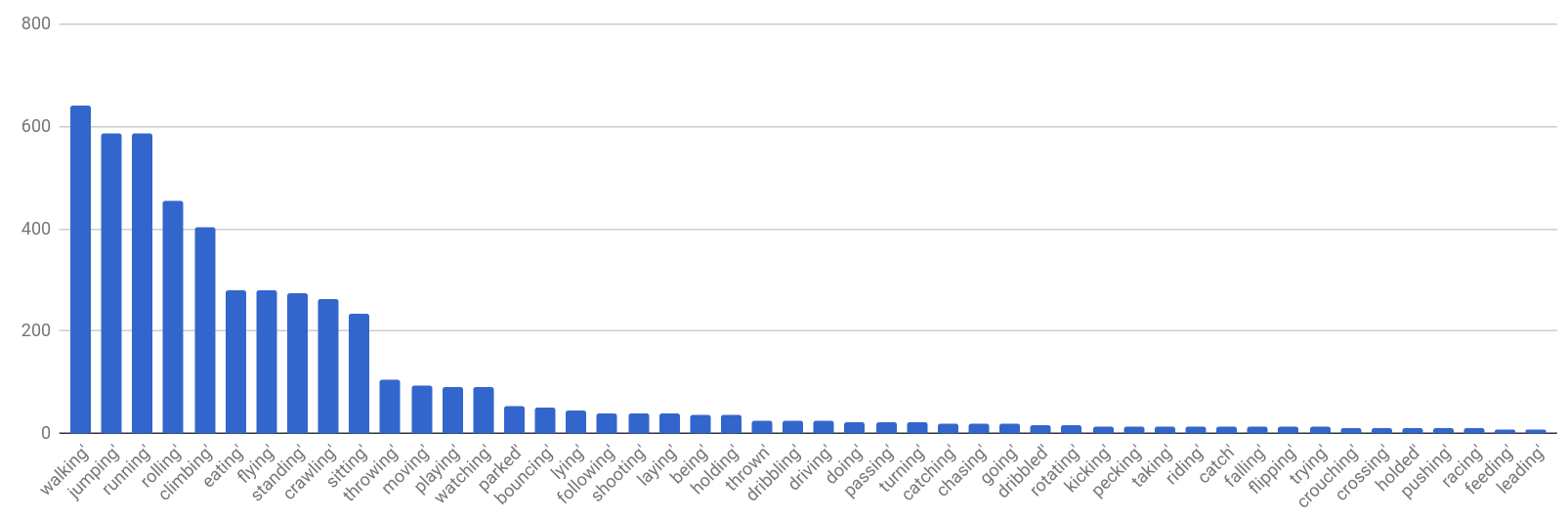} 
    \caption{Most frequent nouns (top) and verbs (bottom) in the A2D Sentences dataset.}
    \label{fig:stats_a2d}
\end{figure*}

\begin{figure*}[t]
    \centering 
    \includegraphics[width=\textwidth,height=0.22\textheight]{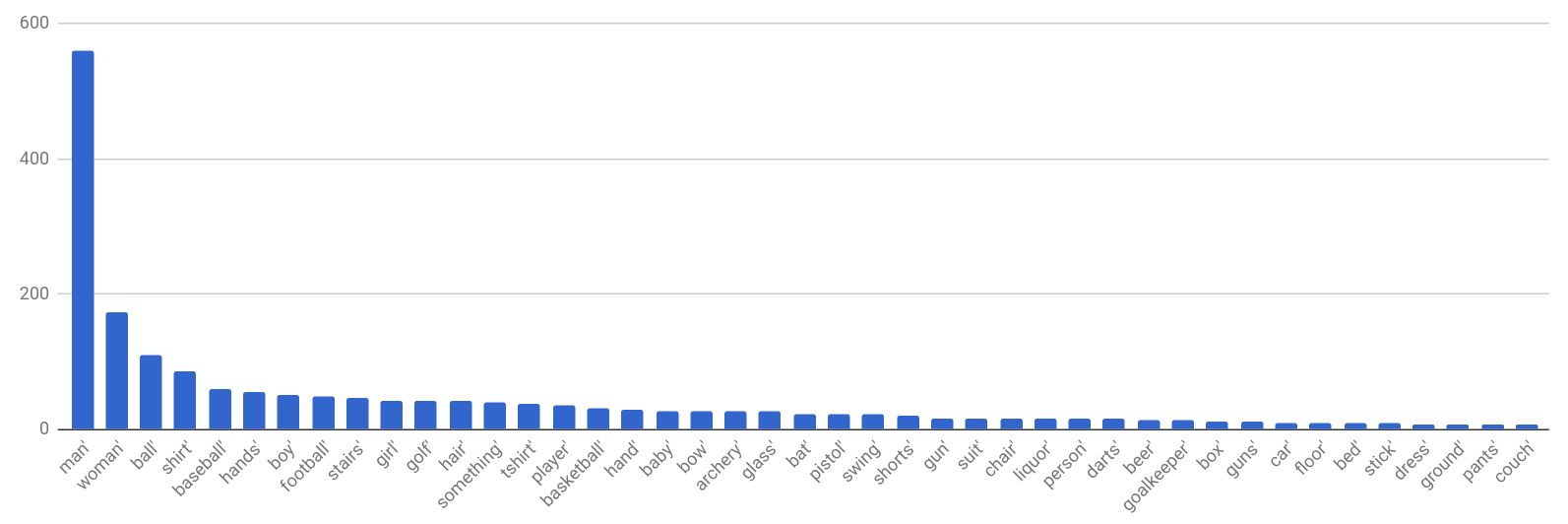}
    
    \includegraphics[width=\textwidth, height=0.22\textheight]{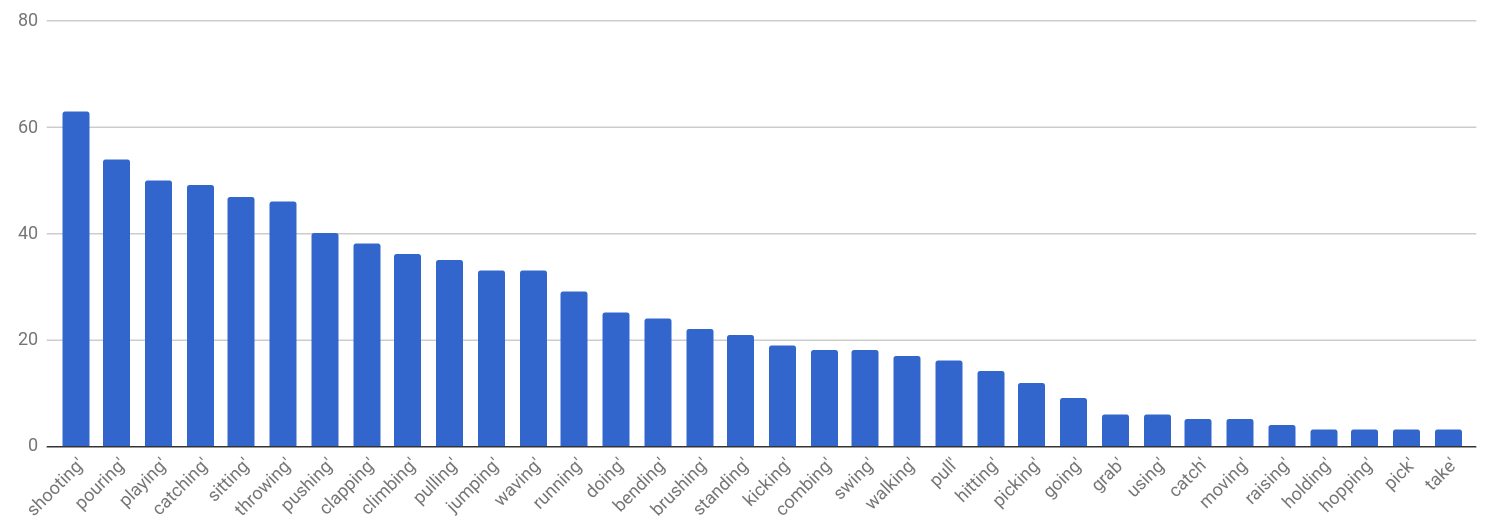} 
    \caption{Most frequent nouns (top) and verbs (bottom) in the J-HMDB Sentences dataset.}
    \label{fig:stats_jhmdb}
\end{figure*}

We show some statistics of the annotated sentences on A2D and J-HMDB datasets. Figure~\ref{fig:stats_a2d} shows the most frequent nouns and verbs in the A2D Sentences dataset. Segmentation from a sentence allows us to distinguish between the fine-grained actors in the same super-category. For example while in the normal A2D dataset~\citeS{supp_xu2015fly} there is a general `adult' category, we annotate fine-grained human actors like \{man, woman, guy, person, girl, boy, ...\} in A2D Sentences. Furthermore, natural language sentences enable us to make use of a richer set of verbs to describe the same type of action,~\eg \{jumping (up and down), bouncing, falling\} all are representative for the action label `jumping' in the regular A2D dataset. Likewise, \{flipping, turning, rolling, rotating\} are representative for the action label `rolling', and \{moving, running, chasing\} are representative for `running'. Figure~\ref{fig:stats_jhmdb} shows the most frequent nouns and verbs in the J-HMDB Sentences dataset. 

\section{Segmentation results on A2D Sentences}\label{sec:res_1_a2d_sentences}
\begin{figure*}[t]
    \centering 
    \includegraphics[width=\textwidth]{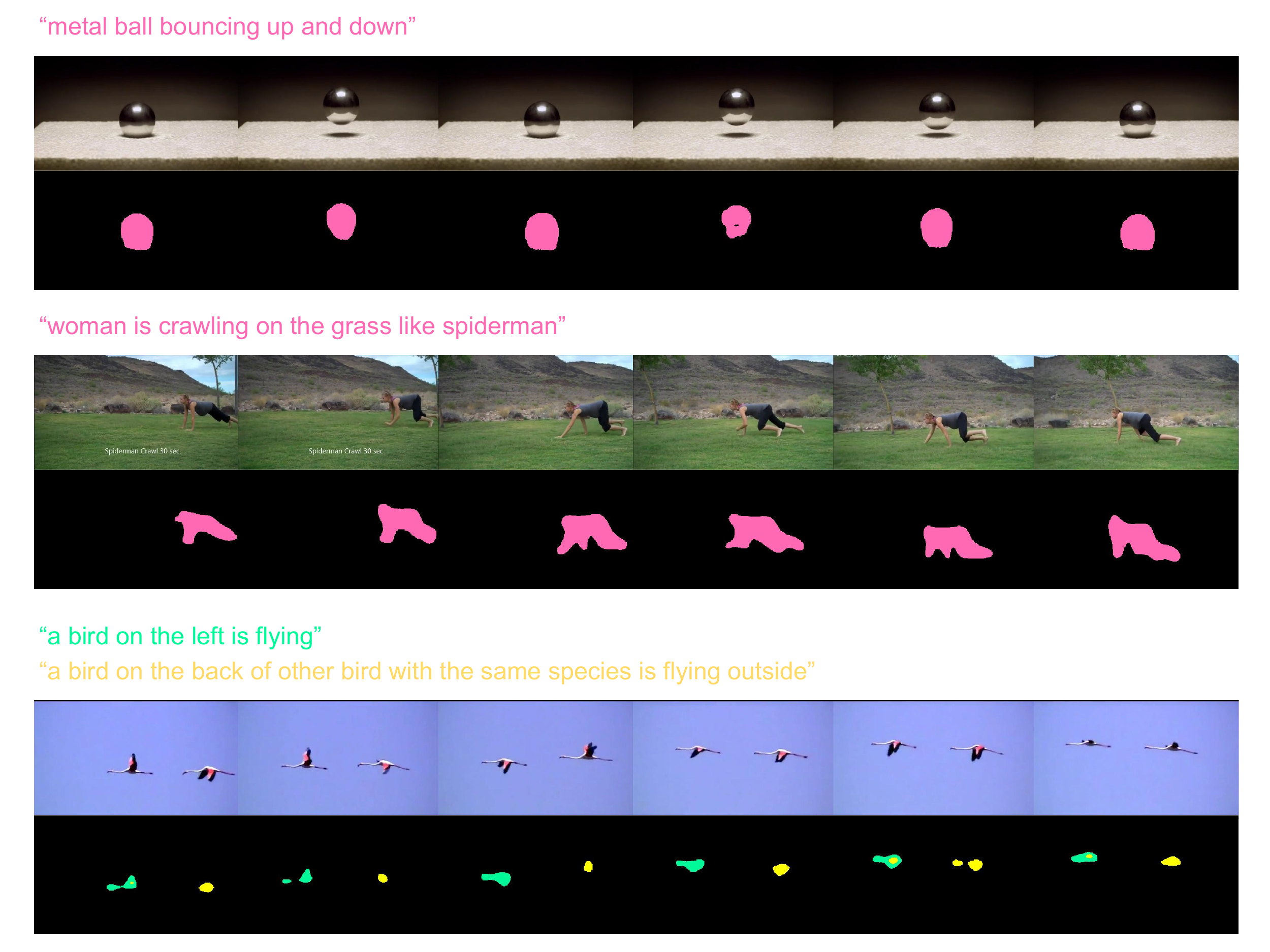} 
    \caption{Visualized segmentation results from our model on A2D Sentences. In all rows we show examples with only one type of actor performing the same action. The first two videos illustrate examples with one single instance while the last video contains two instances. The colored segmentation masks are generated from the sentence with the same color above each video.}
    \label{fig:vis_1}
\end{figure*}

\begin{figure*}[t]
    \centering 
    \includegraphics[width=\textwidth]{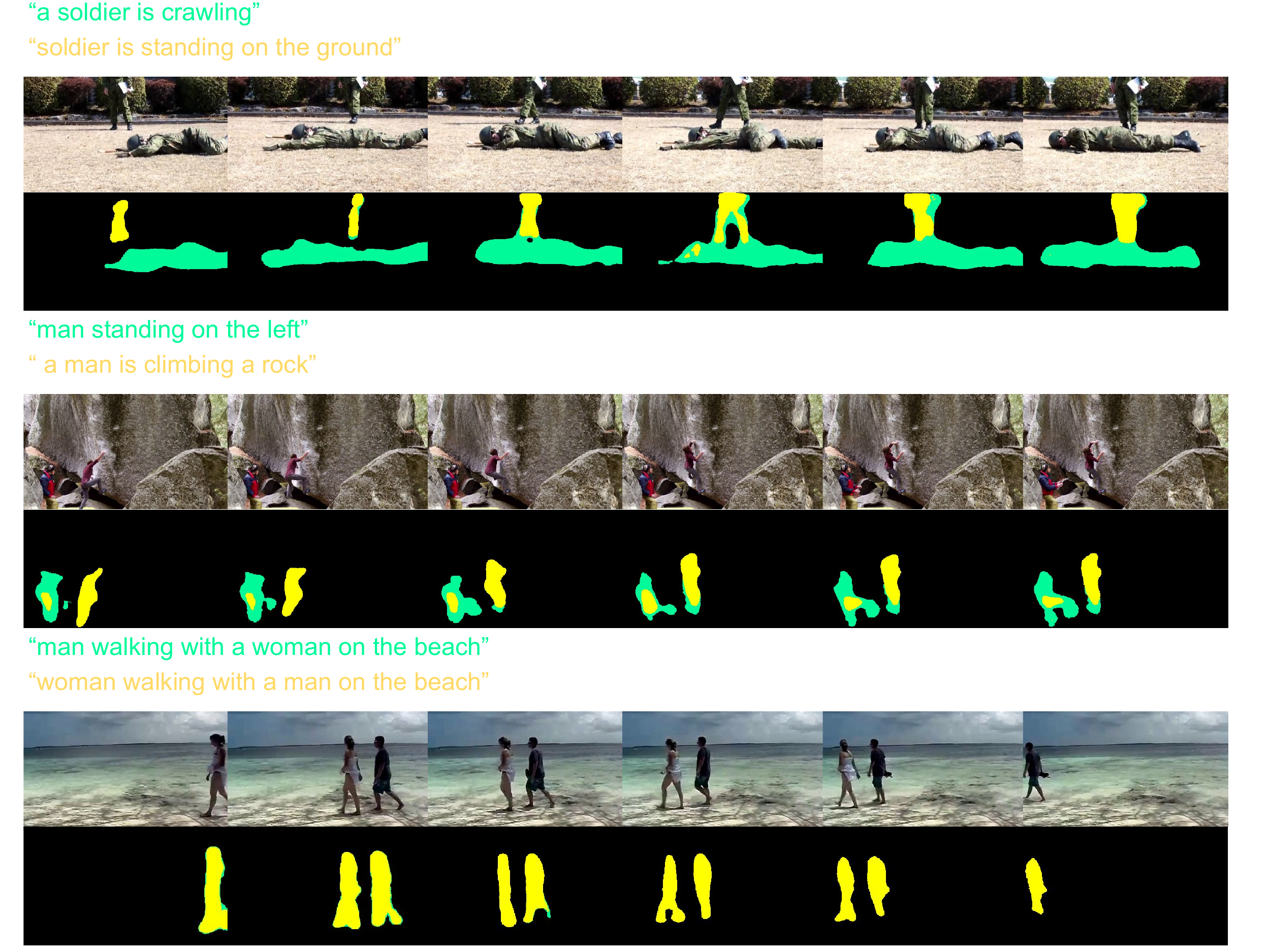} 
    \caption{Visualized segmentation results from our model on A2D Sentences. In the first two rows we show examples with one type of actor performing different actions. The last row illustrates a failure case of our model.  The colored segmentation masks are generated from the sentence with the same color above each video.}
    \label{fig:vis_2}
\end{figure*}

In this section, we visualize more results of the sentence-guided segmentation using our model. Figure~\ref{fig:vis_1} illustrates videos with only one type of actor performing the same action. Our model segments both deformable (\eg, the `woman' in the second video) and non-deformable (\eg, the `ball' in the first video) objects. Also, it can handle reflecting surfaces, indicated by the `ball' example. The third video demonstrates the ability of our model to distinguish instances among the same actor and action type by language cues like the spatial location provided in the sentence descriptions. Figure~\ref{fig:vis_2} illustrates videos showing human actions. While the first two videos prove the ability of our model to recognize different human actions, the last video shows a failure case of our model. The model is asked to segment `man' and `woman' separately, while it segments both.      

\section{Baseline comparison on A2D Sentences}\label{sec:res_2_comp_baseline}
\begin{figure*}[t!]
    \centering 
    \includegraphics[width=\textwidth]{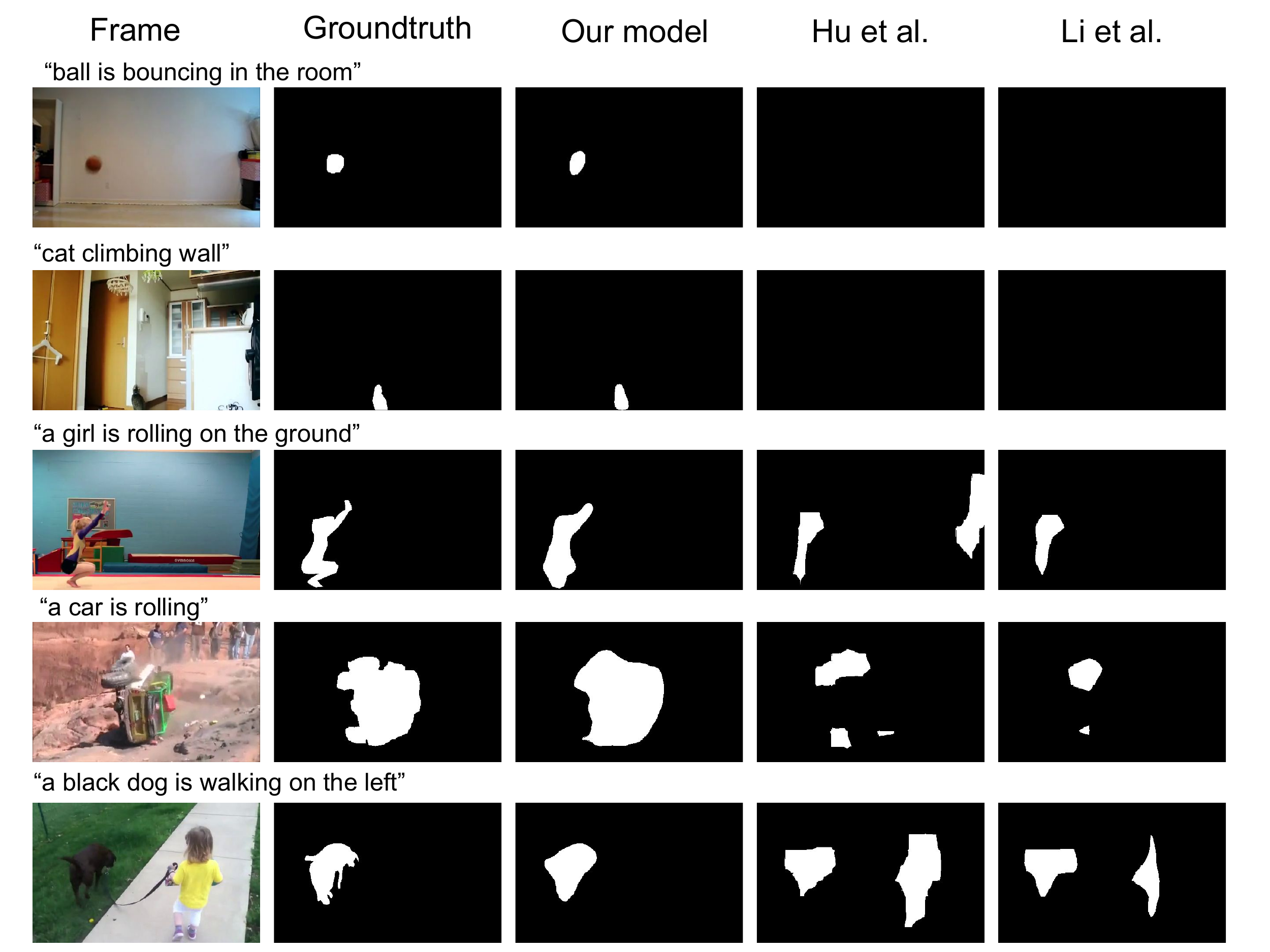} 
    \caption{Qualitative results on A2D Sentences. Columns from left to right are frame to segment, groundtruth segmentation, our model output, output of Hu~\etal and output of Li~\etal. Above each example there is a sentence used as input for all methods describing what to segment in the frame. }
    \label{fig:vis_3}
\end{figure*}

In this section, we show a qualitative comparison of our model with two image-based baselines by Hu~\etal~\citeS{supp_hu2016segmentation} and Li~\etal~\citeS{supp_li2017tracking} in Figure~\ref{fig:vis_3}. The first two rows verify that our model is able to segment relatively small actors, while both baselines struggle. The next two rows demonstrate the better segmentation accuracy of our model in comparison to the baseline models. For example, in the fourth row our model segments the car as a whole, while both baselines segment parts of the car only. In the last row, we illustrate the ability of our model to better distinguish between different types of actors.  

{\small
\bibliographystyleS{ieee}
\bibliographyS{supp_egbib}
}
\end{document}